\documentclass{apa6}
\pdfoutput=1 

\usepackage{url}
\usepackage{xspace}
\usepackage{amsmath}
\usepackage{textcomp}
\usepackage{verbatimbox}

\title{Unobtrusive Low Cost Pupil Size Measurements using Web cameras}

\author{Sergios Petridis, Theodoros Giannakopoulos and Costantine D. Spyropoulos}

\affiliation{National Center for Scientific Research "Demokritos"}

\abstract{Unobtrusive every day health monitoring can be of important use for the elderly population. In particular, pupil size may be a valuable source of information, since, apart from pathological cases, it can reveal the emotional state, the fatigue and the ageing. To allow for unobtrusive monitoring to gain acceptance, one should seek for efficient methods of monitoring using common low-cost hardware. This paper describes a method for monitoring pupil sizes using a common web camera in real time. Our method works by first detecting the face and the eyes area. Subsequently, optimal iris and sclera location and radius, modelled as ellipses, are found using efficient filtering. Finally, the pupil center and radius is estimated by optimal filtering within the area of the iris. Experimental result show both the efficiency and the effectiveness of our approach.
}


\begin{document}

\maketitle

\section{Motivation}

Unobtrusive every day health monitoring can be of important use for the elderly population. In particular, pupil size may be a valuable source of information, since, apart from pathological cases, it can reveal the emotional state, the fatigue and the ageing. To allow for unobtrusive monitoring to gain acceptance, one should seek for efficient methods of monitoring using common low-cost hardware. A low cost camera that monitors the user while in front of a laptop or behind a mirror \cite{poh2011medical} falls into this scenario. Detecting pupils and pupil sizes in this context is of great importance. Namely, pupil sizes may be a valuable source of information, since, apart from pathological cases, it can reveal the emotional state \cite{Partala2003185}, the fatigue \cite{doi:10.1076/0271-3683(200007)2111-ZFT535} and the ageing \cite{winn1994factors} of the subject under monitoring.

Towards this end, this work presents a method for detecting iris and pupils, including both their centers and sizes, from low resolution visible-spectrum images, using a robust unsupervised filter-based approach. Iris detection performance outperforms most state of the art methods compared, and is competitive to few others. With respect to pupil detection, to our knowledge, detecting pupil sizes detection is not reported elsewhere in the related literature. Using a dataset compiled in particular for this purpose, we show  that our method is accurate enough to provide significant information for everyday long-term monitoring.

\section{Relevant Work}

The task of detecting eyes in images or videos is crucial and challenging in many computer vision applications. First, eye detection is a vital component of most face recognition systems, where eyes are used for feature extraction, alignment, face normalization, etc.. In addition, eye tracking is widely used in human computer interaction (gaze tracking). Eye detection systems can be categorized according to the adopted data acquisition method in (a) visible imaging and (b) infrared imaging. According to the first \cite{2001_hausdorffDistance, 2004_eyedetection_projection_functions, 2006_eyeDetectionEdge_asteriadis, 2011_eyeDetection, 2008valentiaccurate, 2004Cristinacceamulti-stage, 2005hamouzfeature}, ambient light reflected from the eye area is captured, hence the task is rather difficult, due to the fact that captured information can contain multiple specular and diffuse components \cite{2006-MS-Dongheng-Li}. On the other hand, infrared-based approaches \cite{2006-MS-Dongheng-Li,2005_StarburstEyeTracking, 2009_GeometrickEyeTracking} manage to eliminate specular reflections and lead to a better and accurate pupil detection. Another discrimination between eye detection approaches is based on the distance of the recording device: (a) head-mounted and (b) remote systems. Needless to say, head-mounted approaches can lead to more 
accurate systems. However, under particular requirements of low cost and low level of obstruction, remote sensing is the only acceptable solution.

\section{Method}

\subsection{Preprpocessing} \label{ssec:preprocessing}

\begin{figure}[h]
  \begin{center}
		\includegraphics[width=7.0cm]{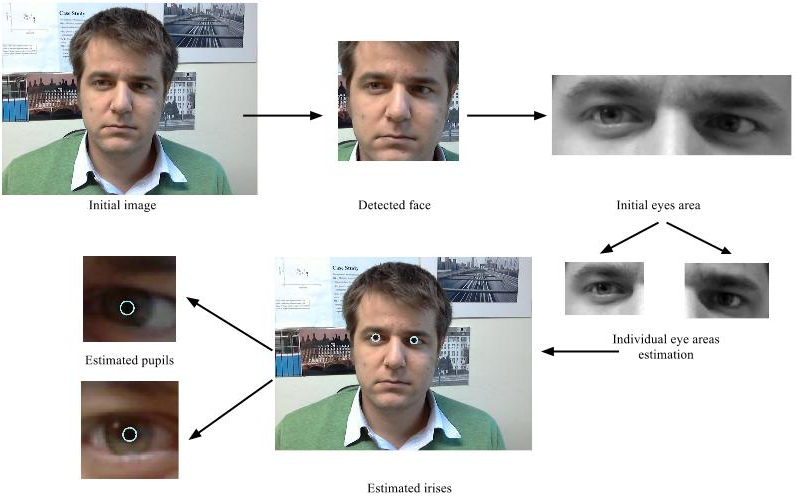}
  \end{center}
  \caption{The overall architecture of the proposed method.}
  \label{fig:method}
\end{figure}

The overall scheme of the proposed method is presented in Figure \ref{fig:method}. 
At a first stage, the face is detected. Face detection is a well studied problem in machine vision \cite{viola2001rapid} and there exist now several commercial tools that achieve high accuracy with high speed. For our purposes, we have used SHORE\textsuperscript{TM}\footnote{SHORE\textsuperscript{TM}:Sophisticated High-speed Object Recognition Engine, Fraunhofer IIS} which achieves face detection at a frame rate greater than 50fps. 

The same engine, also provides directly as a rough estimate of the two eyes area, which we have used to initiate iris and pupil detection. 

\begin{figure}[ht]
  \begin{center}
  \includegraphics[width=0.3\textwidth]{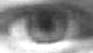}
\end{center}
    \caption{Sample eye area image}
  \label{fig:eye}
\end{figure}

\subsection{Sclera and Iris detection} \label{ssec:iris_detection}

The sclera/iris detection method aims at determining the coordinates $( e^L_x, E^L_y), (e^R_x, e^R_y)$ and the radii $e^L_r, e^R_r$ of the left and right irises, considered as circular disks\footnote{The left (respectively right) eye is denoted by subscript L (respectively R)}. Detection is done independently in each eye and is achieved by maximizing that output of a scoring process, while applying a specialized bank of linear filters parametrized by the radius of the iris, within the rough area of the eye. Therefore, this process results in both the estimation of the center of iris and its radius. The overall score is evaluated as a sum of three scores, based on luminosity, saturation and symmetry, whose definitions are given below.

\begin{figure}[ht]
\begin{center}
  \includegraphics[width=0.3\textwidth]{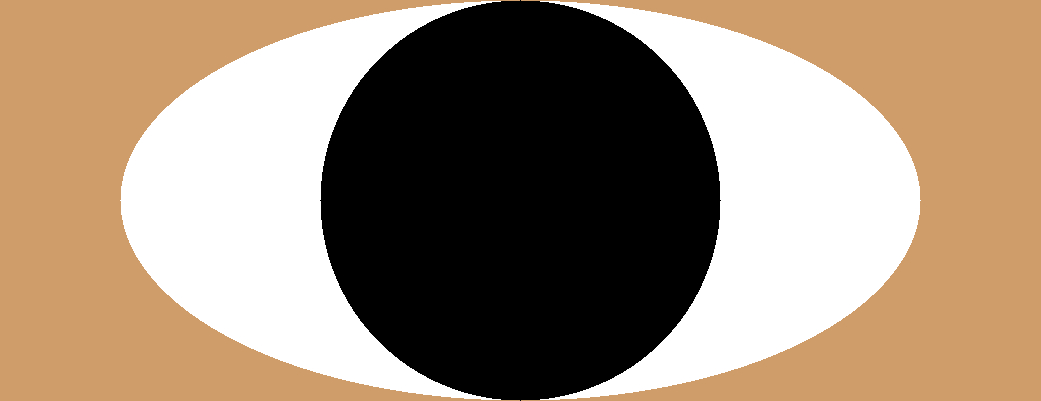}
\end{center}
  \caption{Mask for iris and sclera detection: The figure indicates with different shades the the three regions of the mask. The actual values depend on the criterion used and on the normalisation factor.}
  \label{fig:mask}
\end{figure}

\subsubsection{Detection based on luminosity}

Denoting the luminosity pixels of the eye rough area as $I_L$ and the set of applied masks as $\{M^L_r\}$, the luminosity score is defined as:
\begin{equation}
l( e_x, e_y, e_r ) = I_L[x,y,r] \cdot M^L_r
\end{equation}
where $\cdot$ above denotes element by element multiplication and $I_L[x,y,r]$ is the luminosity values of an image region centered at $(x,y)$ and with size equal to the size of mask $M^L_r$.

The motivation here is that the iris can be located as a region \emph{darker} than the surrounding sclera. 
To that end, we have used a mask with three regions, as depicted in Figure ~\ref{fig:mask}, where the elements of each region all share the same value. In particular:
\begin{description}
\item[iris] a circle centered at the center of mask, where elements have the same \emph{negative} value 
\item[sclera] a region defined as the difference of the above circle and the co-centered ellipse of equal radius along the vertical axis and double radius along the horizontal axis, where elements have the same \emph{positive} value
\item[skin] a region define as the difference of the above ellipse and a co-center rectangular region, where the elements have \emph{zero} value
\end{description}
The mask values are normalized such that they sum up to zero. 

\subsubsection{Detection based on saturation}

To detect the sclera and the iris, it is useful to observe that the sclera is typically much less saturated than both the iris and the surrounding skin. To that end we apply the same method as above using a set of masks $M^S_r$. These masks are similar to the ones used for detection based on luminosity, in that they are composed of the same regions. The difference lie in the value of pixel within each region. Namely, for 
the iris and skin elements share the same \emph{positive} value whereas for the sclera the same \emph{negative} value. As for luminosity, the mask values are normalized such that they sum up to zero. The score based on saturation is then evaluated as
\begin{equation}
s( e_x, e_y, e_r ) = I_S[x,y,r] \cdot M^S_r
\end{equation}
where $I_S$ is the saturation values of the eye rough area considered.

Note that, as a practical approximation, our method actually uses the $V$ channel of the $YUV$ format to approximate saturation.

\subsubsection{Detection based on symmetry}

A further observation to boost the accuracy of sclera and iris detection is that these regions show a significant symmetry. In particular, we have experimentally found that checking for horizontal symmetry inside the iris and sclera regions significantly removes false eye detections. 

In particular, by denoting with superscript H an image region obtain by horizontal flipping, we evaluate the symmetry score as 
\begin{equation}
\begin{split}
h( e_x, e_y, e_r ) &= \bigl(| I_L[x,y,r] - I^H_L[x,y,r] | \\
&+ | I_S[x,y,r] - I^H_S[x,y,r] | \bigr) \cdot M^H_r
\end{split}
\end{equation}
Note that both luminance and saturation values are used here.
The mask $M^H_r$ is of similar structure to $M^L_r$, though with negative elements inside the sclera and iris and zero elements within the skin area.

\subsubsection{Overall Score}

The overall score for each candidate iris center and radius is evaluated as a sum over the respective luminance, saturation and symmetry scores.
\begin{equation}
c( e_x, e_y, e_r ) = l( e_x, e_y, e_r ) + s( e_x, e_y, e_r ) + h( e_x, e_y, e_r ) \label{eq:irisScore}
\end{equation}
and the final choice is made by exhaustively searching over the rough eye area found during preprocessing.
\begin{equation}
( \hat e_x, \hat e_y, \hat e_r ) = \operatorname*{arg\,max}_{e_x, e_y, e_r} c( e_x, e_y, e_r )
\end{equation}

\subsection{Pupil Detection and measurement}

As soon as the disk that defines the iris area for each eye has been estimated, the pupil is detected as a circle within the iris that optimal satisfies a gradient-based criterion. Candidate pupils are circles with center $(p_x,p_y)$ in the close neighborhood of the iris center $(e_x,e_y)$, with radius such that they fall strictly inside the iris area. 

The gradient criterion is evaluated as the difference between the average luminosity of the pixel defining the perimeter of the candidate circle with the average luminosity of the immediate outer pixels of the circle. The greater the difference, the greater the possibility that the circle corresponds to the pupil. Note that the sign of the difference is important here. 

Though this approach is conceptually simple, it has shown that it is quite accurate, even in the presence of light reflections, which may degrade the performance of a non-gradient method.

\subsection{Optimal Frame}

The procedure outlined for detecting the iris and the pupil is repeated for every frame obtained from the camera, for both eyes. Since the ultimate goal is to measure the pupil size, and given that pupil size does not change from frame to frame\footnote{we assume here that lighting conditions stay the same}, it is not needed to measure the pupil on each frame, but rather on a frame where it can be measured with higher confidence. To that end, we describe now a method that evaluates the optimal frame based on which detecting the pupil and measuring its size can be attempted.

Namely, for every frame, we compute an overall \emph{confidence score}, as the product of the following measures:
\begin{itemize}
\item left eye iris detection score
\item right eye iris detection score
\item $e^L_r$ and $e^R_r$ equality based score
\item $e^L_y$ and $e^R_y$ equality based score
\item $p^L_y$ and $p^R_y$ equality based score
\end{itemize}
Regarding the first two items of the above list are directly given through Eq~\ref{eq:irisScore}. The equality based scores are evaluated based on the generic formula
\begin{equation}
s = \frac{| l - r |}{\max \{l,r\}}
\end{equation}
where $l$ (respectively $r$) is a measured obtained from the left (respectively right) eye.

Therefore, as the video is streaming, we evaluate the overall confidence score and compare it to the one that has been obtained so far. In case it is greater, the less confident value is discarded and the new one is kept as the optimal one. In this way, the latest results are always based on the more confident frames. The procedure is repeated until the person under monitoring is stopped from being tracked, or after the end of a predefined time duration. In both case, the confidence score is reset and the procedure is repeated again.

\section{Complexity Analysis}

A main concern in the development of the proposed method has been to keep low the overall complexity. This has been important for two reasons. The first one relates to low resources that the method should be needing. Either running as a background process on a tablet, or as a process on a dedicated hardware, detecting and measuring pupils should take as few resources as possible, given that the same hardware may be hosting other processes too. The second one relates to the speed of execution. Even though pupil size measurement is not critical, and therefore latency is affordable, the need of video recording should nevertheless be avoided to address users\textquotesingle\ privacy concerns. 
Overall, our goal has been to measuring pupils in at least real time given limited processing resources and no storage. 

\subsection{One-pass iteration}

The method that has been described in this paper does achieve this goal. In particular, a significant speed up has been achieved by allowing scores involving iterations over pixels to be computed with a single pass over the respective image region. This has been possible, since all masks share the same structure and therefore scores are simultaneously updated by iterating over the region pixels. Moreover, since mask elements have no more than three values for each mask, the computation requires a number of additions equal to the number of pixel in the region. Multiplications are only constant with respect to image region size.

\subsubsection{Pupil Scores}

In the same direction, we also stress that candidate pupil scores are evaluated within the same iteration. To achieve this, the value of the mask element within the iris is used. In particular, notice that while the sign of the elements can be used to identify that they belong to the iris, the value is of non particular importance when evaluating the luminance and saturation values, since, as noted above, a plain summation of the values within the iris is performed, followed by a normalization using a pre-calculated normalization factor. Therefore, the value of the elements have been used to tag the distance from the mask center, as depicted in Figure~\ref{fig:actualMask}. In this way, while iterating over the mask elements, an array indexed by the distance from the center, progressively accumulates the sum of the values with the same distance from the center. After iterating over all elements, this array will contain, in each element (index), the sum of luminance values for the given index.
{\tiny\begin{verbbox}
0 0 0 0 0 0 0 0 0 0 0 0 0 0 0 0 0 0 0 8 0 0 0 0 0 0 0 0 0 0 0 0 0 0 0 0 0 0 0 
0 0 0 0 0 0 0 0 0 0 0 0 - - - - 8 8 8 7 8 8 8 - - - - 0 0 0 0 0 0 0 0 0 0 0 0 
0 0 0 0 0 0 0 0 0 0 - - - - - 8 7 7 7 6 7 7 7 8 - - - - - 0 0 0 0 0 0 0 0 0 0 
0 0 0 0 0 0 0 0 - - - - - - 8 7 6 6 6 5 6 6 6 7 8 - - - - - - 0 0 0 0 0 0 0 0 
0 0 0 0 0 0 0 - - - - - - 8 7 6 6 5 5 4 5 5 6 6 7 8 - - - - - - 0 0 0 0 0 0 0 
0 0 0 0 0 0 - - - - - - - 8 7 6 5 4 4 3 4 4 5 6 7 8 - - - - - - - 0 0 0 0 0 0 
0 0 0 0 0 0 - - - - - - - 8 7 6 5 4 3 2 3 4 5 6 7 8 - - - - - - - 0 0 0 0 0 0 
0 0 0 0 0 - - - - - - - 8 7 6 5 4 3 2 1 2 3 4 5 6 7 8 - - - - - - - 0 0 0 0 0 
0 0 0 0 0 0 - - - - - - - 8 7 6 5 4 3 2 3 4 5 6 7 8 - - - - - - - 0 0 0 0 0 0 
0 0 0 0 0 0 - - - - - - - 8 7 6 5 4 4 3 4 4 5 6 7 8 - - - - - - - 0 0 0 0 0 0 
0 0 0 0 0 0 0 - - - - - - 8 7 6 6 5 5 4 5 5 6 6 7 8 - - - - - - 0 0 0 0 0 0 0 
0 0 0 0 0 0 0 0 - - - - - - 8 7 6 6 6 5 6 6 6 7 8 - - - - - - 0 0 0 0 0 0 0 0 
0 0 0 0 0 0 0 0 0 0 - - - - - 8 7 7 7 6 7 7 7 8 - - - - - 0 0 0 0 0 0 0 0 0 0 
0 0 0 0 0 0 0 0 0 0 0 0 - - - - 8 8 8 7 8 8 8 - - - - 0 0 0 0 0 0 0 0 0 0 0 0 
0 0 0 0 0 0 0 0 0 0 0 0 0 0 0 0 0 0 0 8 0 0 0 0 0 0 0 0 0 0 0 0 0 0 0 0 0 0 0 
\end{verbbox}
}
\begin{figure}
\begin{center}
\theverbbox
\end{center}
\caption{Actual mask values for mask of iris radius 8. Positive values denote iris or pupil, the minus sign denotes (actually -1) denotes the sclear and the zero value the skin.}\label{fig:actualMask}
\end{figure}

\subsubsection{Parallelization}

Note also that our method can be graciously parallelized in many cores --- one instruction processing architecture which would allow a further speedup on the execution. Such a solution is highly desirable in cases which one wishes to make the most out of a dedicated hardware including both CPU and GPU. Actually, the authors are currently implementing the method using the CUDA programming language, such that it can be executed on a low energy consumption nettop (Zotac Z-BOX ID84 PLUS) featuring a Intel Atom D2550 1.86 GHz Dual-Core CPU and NVIDIA GeForce GT 520M (512 MB) GPU. 

\subsubsection{Tuning the frame rate}

We further notice that in case the method needs to be implemented in a lower processing capabilities hardware, real time analysis can be guaranteed by lowering down the video frame rate per second. Of course, in this case, a lower accuracy may be noticed, given that frames containing clearer pupil sizes may have been missed. The frame rate achieved in the Zotac Z-BOX ID84 PLUS using only CPU has been 5 frames per second, whereas a frame rate above 30 has been achieve for a PC feature an Intel Core i5-2500 CPU @ 3.30GHz.

\section{Results}
Our evaluation had two purposes. First, to evaluate the performance of the iris detection module. In this case, we are only interested in estimating the iris center (not its size), since most of the related publicly available datasets only have center annotations (e.g., \cite{bioID}). Second, to evaluate the performance of pupil detection. Here, we are interested in estimating the exact pupil area (center and radius). Towards this end, we have built a dataset with pupil-related annotations.

\subsection{Iris center localization performance} \label{ssec:eyeCenter}

The proposed method has been evaluated against nine state of the art methods. Two of them are provided by the MATLAB Vision Toolbox \cite{cartMatlab, CastrillonDGH07}, namely (a) CART  \cite{cartMatlab} and (b) HAAR  \cite{CastrillonDGH07}, and allowed an in depth comparison using several performance measures. The others have been compared using reported results in \cite{2001_hausdorffDistance, 2004_eyedetection_projection_functions, 2006_eyeDetectionEdge_asteriadis, 2011_eyeDetection, 2008valentiaccurate, 2004Cristinacceamulti-stage, 2005hamouzfeature}.
Furthermore, a "Rough" estimation has been used as baseline based on setting the estimated iris center as the center of the initial individual eye areas, which are extracted as explained in Section \ref{ssec:preprocessing}. In some cases only the results of MATLAB-related methods are shown, since these are reproducible, while all compared methods are only shown for the case of Table \ref{tbl:results_tol}.
We compared the methods against the widely used BioID dataset \cite{bioID}, using all available samples. BioID test cases include a larger variety of illuminations conditions. 

The performance measures involved in this evaluation are defined as follows. First, let $d_l$ (respectively $d_r$ be the euclidean distance between the detected and manually annotated left (respectively right) iris centers. Also, let $d_{lr}$ be the distance between the manually annotated left and right iris centers. The \emph{relative errors} for the two detected irises are evaluated as
\( e_l = \frac{d_l}{d_{lr}} \text{ and } e_r = \frac{d_r}{d_{lr}} \)
whereas the relative error over both eyes as $e = (e_l + e_r ) / 2$. The respective error measures over the dataset are naturally defined as average errors over all dataset samples: $E_l = \frac{1}{N}\sum_{i=1}^{N} e_l(i)$, $E_r = \frac{1}{N}\sum_{i=1}^{N} e_r(i)$ and $E = \frac{1}{2}(E_l + E_r)$, where $N$ is the total number of samples in the testing set. 
Table \ref{tbl:results_avg} shows the average iris detection errors for the compared MATLAB-related methods. The proposed method outperforms all compared methods.

\begin{table}[t]
\begin{center}
  \begin{tabular}{ l  l l l}
    \toprule
    Method     &  $E_l$ & $E_r$ & $E$ 	\\ \midrule
    HAAR       &  0.052 	    &  0.040  & 0.046 		\\ 
    CART       &  0.060 	    &  0.057  & 0.058 		\\ 
    Rough      &  0.054 	    &  0.053  & 0.053 		\\ 
    \bf{Proposed}   &  \bf 0.035 	    &  \bf 0.021  & \bf 0.028 		\\ 
  \bottomrule
  \end{tabular}
  \end{center}
  \caption{Iris Detection Average Relative Error (left, right and overall) results (\emph{MATLAB-related methods compared})}
  \label{tbl:results_avg}
\end{table}

Furthermore, we have compared our method against an tolerance-based accuracy measure widely used in the literature \cite{2001_hausdorffDistance, 2004_eyedetection_projection_functions, 2006_eyeDetectionEdge_asteriadis, 2011_eyeDetection, 2008valentiaccurate, 2004Cristinacceamulti-stage, 2005hamouzfeature}. Namely, given an error tolerance $T$, an eye detection result is considered as successful if both errors $e_l$ and $e_r$ are less than $T$:
\begin{equation}
 A_T = \frac{\sum_{i: max(e_l(i), e_r(i)) \leq T} 1 }{N}
\end{equation} 
Using this measure, it has been possible to compare the proposed method against \cite{2001_hausdorffDistance, 2004_eyedetection_projection_functions, 2006_eyeDetectionEdge_asteriadis, 2011_eyeDetection, 2008valentiaccurate, 2004Cristinacceamulti-stage, 2005hamouzfeature} as well, using the corresponding reported results.
A typical value for the threshold is $T=0.25$, because it corresponds to an accuracy of about half the width of an eye in the image, while $T=0.1$ has also been used \cite{2004_eyedetection_projection_functions, bioID, 2006_eyeDetectionEdge_asteriadis}. In Table \ref{tbl:results_tol} the tolerance-based accuracy for three different tolerance thresholds is presented. The proposed method outperforms most of the compared methods, except for the case of $T=0.05$. 

\begin{table}[t]
\begin{center}
  \begin{tabular}{ l  l l l }
    \toprule
               &  \multicolumn{3}{c}{$T$} 	\\ \cmidrule{2-4}
    Method     &  0.05    &  0.1 & 0.25 	\\ \midrule	    
    HAAR       &  22 	  &  75  & 98 		\\ 
    CART       &  8 	  &  68  & 99 		\\ 
    Rough      &  15 	  &  68  & 98 		\\ 
    \cite{2001_hausdorffDistance} 			& 40  & 80 & 91 \\
    \cite{2004_eyedetection_projection_functions}	& -   & -  & 95 \\		    
    \cite{2006_eyeDetectionEdge_asteriadis}		& 50  & 82 & 98 \\		    
    \cite{2011_eyeDetection}				& 45  & 85 & 95 \\
    \cite{2008valentiaccurate}				& 84  & 91 & 99 \\
    \cite{2004Cristinacceamulti-stage}			& 56  & 96 & 98 \\
    \cite{2005hamouzfeature}				& 59  & 77 & 93 \\    
    \bf{Proposed}   					& 47  & 92 & 99 \\ 
    \bottomrule
  \end{tabular}
\end{center}  
  \caption{Iris detection tolerance-based accuracy results for three different tolerance values (\emph{all methods compared})}
  \label{tbl:results_tol}
\end{table}

\subsection{Pupil size estimation performance} \label{ssec:pupilPerformance}

Pupils size estimation evaluation needs a different setting from the one described above.
To begin with, annotations of the pupil's whole area, not just its center, is needed. To our knowledge, there is no available dataset with such annotations. Therefore, we have built a manually annotated dataset of pupil area. Three humans annotated the irises and pupils, to estimate an inter-annotator agreement level. Final pupil annotations have been considered s the average ares of all annotators. The compiled dataset consists of 50 1280x960 resolution face images of 10 different humans.

With respect to the performance measures, we have used the recall, precision and F1 measures over the pupil areas. In particular, let $A_a$ be the area of the manually annotated pupil, $A_e$ be the area of the estimated pupil and $A_c$ be the area of their intersection. The recall, precision and $F_1$ rates are defined as $R = \frac{A_c}{A_a}$, $P = \frac{A_c}{A_e}$ and $F1 = \frac{2\cdot R \cdot P}{R + P}$ respectively. 

These same measures have also been used for evaluating the inter-annotator agreement. The results of the evaluation are displayed in Table \ref{tbl:results_pupil}. Note that the inter-annotation agreement scores are relatively low, revealing the difficulty even for a human in annotating the pupil area, especially for dark and brown eyes. 

In terms of average $F_1$ measure, the proposed pupil detection method is $12\%$ less accurate than the human annotation performance. In terms of pupil diameter size estimation, our method achieves on average $85\%$ accuracy. This means that for an average 6mm pupil diameter, average error is 0.9mm, which is much lower than the threshold of 2mm indicated for significant pupil differentiations \cite{Partala2003185,winn1994factors}. 

A specific benchmark dataset containing such cases would nevertheless be needed to explicitly verify this conclusion. 

\begin{table}[t]
\begin{center}
  \begin{tabular}{ l  l l l }    \toprule
		           					&  Precision & Recall & F1 \\ \midrule	    
    i-a.a           			 	&  82 	    &  84  & 79 		\\     
    Proposed Method 			  	&  66		&  68   & 67    \\ 
    \bottomrule
  \end{tabular}
\end{center}  
  \caption{Pupil detection performance and comparison to inter-annotator agreement (i-a.a).}
  \label{tbl:results_pupil}
\end{table}

\section{Discussion}

A method for iris and pupil detection, including their sizes, has been presented, based on a robust unsupervised recursive filtering technique. Evaluation on iris center detection has shown that the proposed method outperforms most of related algorithms. Pupil size estimation was evaluated on a separate dataset which also contains annotations regarding not only pupils position but their sizes as well. The final pupil detection performance results showed that the proposed method's accuracy is accurate enough to be considered as a low cot pupillometry for long-term monitoring. Further results on video should be presented to show the utility of the confidence of the result based on frame sequences.

\section*{Acknowledgments}
The research leading to these results has been funded by the European
Union's Seventh Framework Programme (FP7/2007-2013) under grant agree-
ment no 288532. For more details, please see \protect\url{http://www.usefil.eu.}

\bibliographystyle{apalike}
\bibliography{low_cost_pupilometry}

\end{document}